\def\year{2021}\relax
\documentclass{article}
\usepackage{ijcai21}

\usepackage{times}
\usepackage{soul}

\usepackage{helvet} % DO NOT CHANGE THIS
\usepackage{courier}  % DO NOT CHANGE THIS
\usepackage[hyphens]{url}  % DO NOT CHANGE THIS
\usepackage{graphicx} % DO NOT CHANGE THIS
\usepackage{caption} % DO NOT CHANGE THIS AND DO NOT ADD ANY OPTIONS TO IT
\usepackage{xcolor}

\newcommand{\argmax}{argmax}
\newcommand{\argmin}{argmin}

\usepackage{amsmath}
\usepackage{amsfonts}
\usepackage{amssymb}

\usepackage{amsthm}

\usepackage{caption}
\usepackage{subcaption}

\theoremstyle{definition}
\newtheorem{definition}{Definition}[section]
\usepackage{eso-pic}% http://ctan.org/pkg/eso-pic

\usepackage{lineno}
\title{
A Unifying Bayesian Formulation of Measures of Interpretability \\ in Human-AI Interaction\thanks{To appear in IJCAI-2021}
}

\author{
Sarath Sreedharan$^1$ $\cdot$ 
Anagha Kulkarni$^1$ $\cdot$ 
\Large {David E. Smith $\cdot$ Subbarao Kambhampati$^1$}\\[1ex]
\textnormal{$^1$Arizona State University}
}

\begin{document}
% \AddToShipoutPictureFG*{%
%   \AtPageUpperLeft{%
%     \hspace{580pt}%
%     \raisebox{-26pt}{%
%       \makebox[3pt][r]{\textbf{Paper under review--please do not circulate}}
% }}
% }%

\maketitle

\begin{abstract}
Existing approaches for generating human-aware agent behaviors have considered different measures of interpretability in isolation. 
Further, these measures have been studied under differing assumptions,
%have generally studied these properties under differing assumptions, using distinct mathematical tools. 
thus precluding the possibility of designing a single framework that captures these measures under the same assumptions.
In this paper, we present a unifying Bayesian framework that models a human observer's evolving beliefs about an agent and thereby define the problem of {\em Generalized Human-Aware Planning}.
We will show that the definitions of interpretability measures like explicability, legibility and predictability from the prior literature fall out as special cases of our general framework.
%capture reasoning about all three of the most popular interpretability measures, namely, {\em legibility, predictability, and explicability} in a single setting general enough to meaningfully exhibit all of the measures. 
%By doing this, we provide a single mathematical account for existing interpretability work and also discuss how these measures can be mapped to more general settings.  
Through this framework, we also bring a previously ignored fact to light that the human-robot interactions are in effect open-world problems, particularly as a result of modeling the human's beliefs over the agent. Since the human may not only hold beliefs unknown to the agent but may also form new hypotheses about the agent when presented with novel or unexpected behaviors. 
%As we will see, this is central to understanding and framing explicability in a Bayesian framework compatible with other measures.
\end{abstract}

\section{Introduction}
A crucial aspect of the design of human-aware AI systems is the synthesis of interpretable behavior \cite{gunning2019darpa,langley2017explainable}. 
Existing works in this direction \cite{chakraborti2019explicability} 
explore behaviors that instigate a desired change in the human's mental state 
or conform with her current mental state so as to not require explicit 
communication. 
Three distinct notions of interpretability 
can be seen in prior work: 
{\em legibility} -- the agent signaling its objectives through behavior (c.f. \cite{dragan2013legibility,dragan2017robot,kulkarni2019unified,kulkarni2019signaling,macnally2018action,dragan2013generating,SZ:MZaamas20});
{\em explicability} -- agent behavior that conforms with the human's expectation  (c.f. \cite{zhang2017plan,kulkarni2019explicable,kulkarni2020designing,balancing}); and 
{\em predictability} -- agent behavior that is easier to anticipate (c.f. \cite{fisac2020generating,dragan2017robot,dragan2015effects,dragan2013legibility}). 
These notions of interpretability can each improve human-AI 
collaborations along different dimensions. 
If you know your agent's objectives (legibility) and can anticipate 
its future behavior (predictability), 
you can plan around it or even exploit it; while in conforming to your 
expectations (explicability), it can avoid surprising you which would 
adversely affect the fluency of collaboration. 

\begin{figure}[pt!]
\centering
\includegraphics[width=0.75\columnwidth]{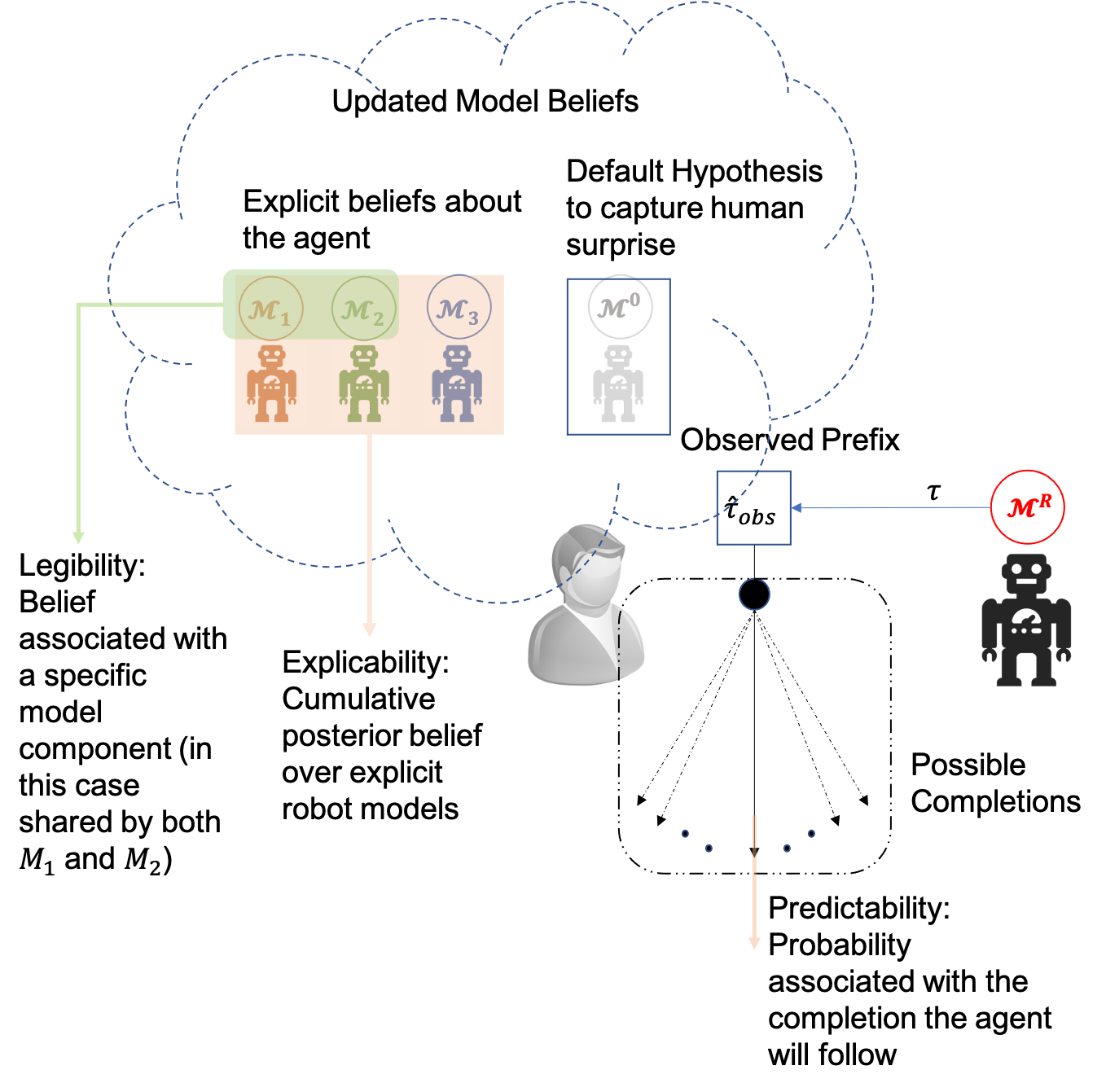}
\caption{
An overview of our unifying framework. The human holds multiple hypotheses about the agent and she uses the observed behavioral prefix to update her beliefs about the model. Each of the interpretability measure optimizes for specific inferential outcomes in this framework. 
}
\vspace{-15pt}
\label{fig:overview}
\end{figure}

While works like \cite{chakraborti2019explicability} provide overarching taxonomies and establishing equivalence between works done in this space, they do not generalize the setting itself to allow for a single unifying framework for these measures.
%we are unaware of any unifying mathematical framework that can account for all of these measures. 
%Though there is a general consensus that these measures are just special cases of more general collaborative strategies agents may engage in, we are unaware of any efforts to investigate such measures (and their potential generalizations) in a single unified setting.
%} 
Two major roadblocks to such a unification are: 1) the measures are defined under competing assumptions and
%-- prior formulations define these measures under incompatible settings. For instance, legibility, which is defined for cases where the observer has multiple hypotheses about the agent (one of which is guaranteed to be the true underlying hypothesis) would not make sense in traditional predictability or explicability settings where the observers assume a single hypothesis about the agent 
2) different frameworks are used to reason about the human's beliefs (which is central to defining these measures).
%the formulations are disparate -- e.g. while  predictability and legibility have been formulated in a Bayesian frameworks, explicability has been formulated using plan distances. 
In this work, we present a single Bayesian framework 
that capture human's reasoning over a distribution of models they ascribe to the agent and use it to
%to capture human's reasoning and use it to
%capture the evolution of an observer's belief about the agent in that setting. We will use this framework to 
define the {\em Generalized Human-Aware Planning} problem. We will show how measures studied in prior literature can be seen as special cases of our unifying framework. Furthermore, our Bayesian formulation of explicability reveals an important dimension of the human-aware planning problem that to the best of our knowledge has not been explicitly studied before -- relationship between explicability and open-world beliefs. That is, by identifying inexplicable behavior, the human identifies that her belief about the agent was incorrect and that the agent's behavior is stemming from an unknown model. 
%the process of a human identifying inexplicable behavior consists of identifying that their beliefs about the agent were incorrect and that the behavior is being generated by a previously unknown model. 
Our unifying framework accommodates this by considering an additional hypothesis that the human's belief about the agent may be wrong. %As we will see, this framework in fact subsumes existing formulations of explicability.
The summary of our contributions is as follows:\looseness=-1
\begin{enumerate}
\item We formulate a Bayesian framework to capture a human observer's reasoning about the agent in terms of a distribution over models and within it define the Generalized Human-Aware Planning Problem.
\item We show that this single unifying framework can be specialized to existing interpretability measures under the original assumptions made by those works.
\begin{itemize}
\item[-] 
The ability to model an ``unknown'' model is critical to the unification of these competing measures.
\item[-] 
The unification further generalizes these measures.
\end{itemize}
\end{enumerate}

%In the discussion section, we also provide a sketch of how this new formulation can be used for planning, and further discuss how we can leverage communication to boost these interpretability properties.

\section{Background}
\label{background}
%Through most of the discussion we will be agnostic to the specific models used to represent the agent. We will also use the term model in a general sense to include information not only about the actions that the agents are capable of doing and their effects on the world but to also include information on the current state of the world, the reward/cost model and any goal states associated with the problem.
In this paper, we will be agnostic to specific planning formulations or representations when discussing the agent's model. Instead, we will use the term ``model" in a general sense to not only include information about agent actions and transition functions, but also their reward/cost function, goals and initial state.
We will assume that the model can be parameterized and use $\theta_i(\mathcal{M})$ to characterize the value of a parameter $\theta_i$ for the model $\mathcal{M}$. %and $\Theta(\mathcal{M})$ to represent the set of all parameter values that uniquely identifies a model.

Since we are interested in cases where a human is observing an agent acting in the world, we will mainly focus on agent behavioral traces (instead of plans or policies).
%Specifically, a 
A behavior trace in this context will consist of a sequence of state, action pairs $\tau$.
%, which we refer to as a {\em trace}.  
The likelihood of the sequence given a model will take the form $P_{\ell}: \mathbf{M}\times \mathcal{T} \rightarrow [0,1]$, where $\mathbf{M}$ is the space of possible models and $\mathcal{T}$ the set of  behavioral traces the agent can generate. 
While we will try to be agnostic to likelihood functions, a fairly common approach \cite{fisac2018probabilistically,baker2007goal}
is a noisy rational model based on the Boltzmann distribution:
$P_{\ell}(M, \tau) \propto e^{-\beta \times C(\tau)}$. Where $C(\tau)$ is the cost of the behavior and  and $\beta \in \mathbb{R}^+$  
is a parameter that reflects level of perceived determinism in the agent's choice of plans \cite{baker2009action}. Note that in our case, a likelihood function captures both the human's expectations about the agent's computational capabilities and their own cognitive limitations. Thus noisy-rational models like the one mentioned above are particularly useful in our scenario. For example, by setting a low $\beta$ value we could possibly capture the fact that the observer may not be able to correctly differentiate between strategies of relatively similar costs. 
%Though we could also employ more sophisticated computational models in its place to capture the observer's expectations more closely.

For the human-aware scenario, we are dealing with two different models \cite{dragan2017robot,chakraborti2018foundations,reddy2018you}: the model that is driving the agent behavior (denoted $\mathcal{M}^R$) and the human's belief $\mathcal{M}^R_h$ about it.
We make no assumptions about whether these two models are represented using equivalent representational schemes or use the same likelihood functions. 
{\em This setup assumes that while the human may have expectations about the agent's model, she may have no expectation about its ability 
to model her. Thus she isn't actively expecting the agent to mold its behavior to what she thinks the agent knows about her, thereby avoiding additional nesting of beliefs.} 

\subsubsection{Running Example}
\label{example}

\begin{figure}[tbp!]
\centering
\includegraphics[width=0.8\columnwidth]{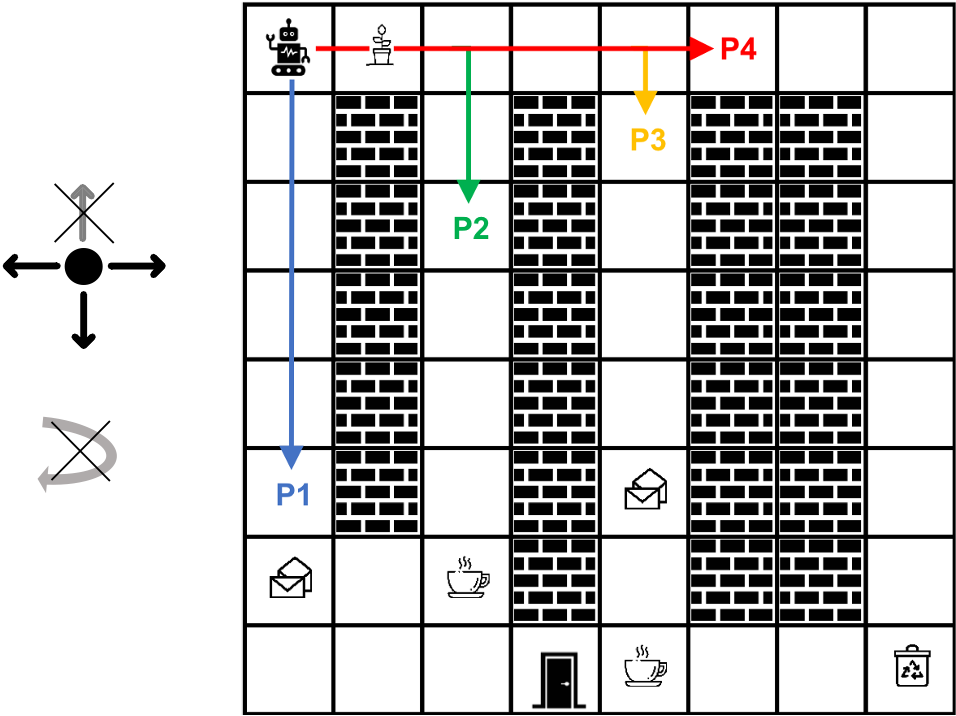}
\caption{
An illustration to show different interpretable behaviors. Here the agent only moves in three directions: down, 
left and right; and it does not revisit a cell.
}
\vspace{-15pt}
\label{fig:office-1}
\end{figure}

In our running example, we will consider a robotic office assistant (Figure \ref{fig:office-1}), 
that can perform various repetitive tasks in the office, including
picking up and delivering various objects to employees,
emptying trash cans, and so on.
%Unlike a standard gridworld scenario, 
Further, we will assume it can only move in three directions: down, 
left and right; and that it can not revisit a cell. These restrictions allow us to control the set of possible completions of a given plan prefix.
You, as the floor manager, are tasked with observing the agent 
and making sure it is working properly. 
Given your previous experience, 
you have come to form expectations about its 
capabilities and its tasks: 
e.g. you may think that the goals of the agent are 
either to deliver coffee or to deliver mail to a room
(represented by the door), 
though you know there may be other possibilities 
that you have not considered. 
Unbeknownst to you, the agent is trying to deliver coffee and it needs to do this while keeping in mind your beliefs about it.
This scenario is particularly designed to accommodate the considerations made by prior works on interpretable behaviors. Throughout this paper, we will revisit this example to show different behaviors.
%to see the kind of behaviors that will be prescribed by our formulation for different requirements.

\section{A Unified Framework}

\begin{figure}
\centering
\includegraphics[width=0.6\columnwidth]{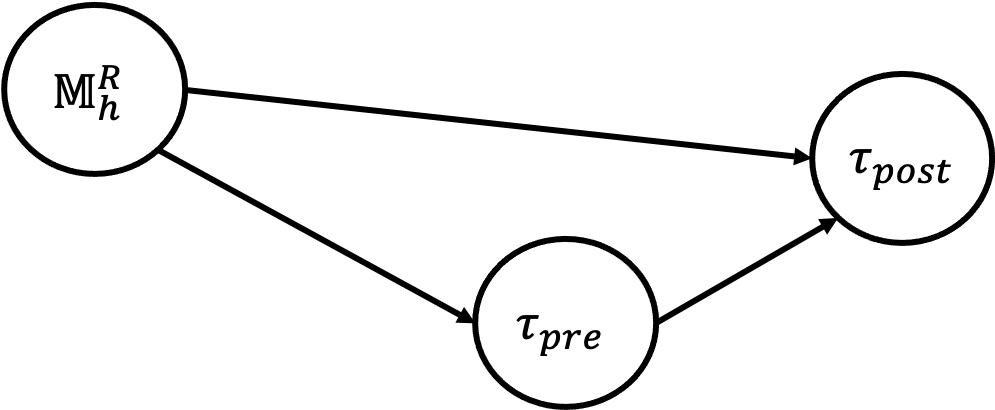}
\caption{Graphical representation of the human's model.}
\label{fig:model}
\vspace{-10pt}
\end{figure}

The ability to anticipate and shape a human's beliefs about the agent, is a central requirement for any successful human-aware agent. So we start with a framework to capture the human's reasoning about their beliefs about the agent. In particular, we will adopt a Bayesian model of the human's reasoning process (Figure \ref{fig:model}). This is motivated by both the popularity of such models in previous works in observer modeling and existing evidence to suggest that people do engage in Bayesian reasoning \cite{l2008bayesian}. 
The node $\mathbb{M}^R_h$ represents possible models the human thinks 
the agent can have, 
 $\tau_{pre}$ corresponds to the behavior prefix that they observed, 
and $\tau_{post}$ corresponds to possible completions of the 
prefix.\footnote{
In this paper, we focus on quantifying these measures 
for one shot or episodic interactions only, rather than longitudinal ones.
In Section \ref{conclusion}, we discuss more about longitudinal interactions.
}

In addition to explicit models that the human thinks are possible
for the agent,
we also allow for the possibility that the human may realize that she in fact doesn't know the exact agent model.
That is her previously held beliefs about the agent may not be sufficient to explain or justify the observed behavior.
We incorporate this assumption by adding a special model 
$\mathcal{M}^0$ to the set of models in $\mathbb{M}^R_h$, that corresponds to the hypothesis that the agent model is not one of the models that the human expects.This allows for open-world reasoning since the human can form additional hypotheses about the robot and is not limited by the explicit set she originally has.
This strategy of introducing a specific hypothesis that corresponds to a previously unexpected entity has been commonly used to model scenarios where there is a possibility of novel or previously unknown event happening (c.f. \cite{pred-unpred}). 
We represent $\mathcal{M}^0$ using a high entropy model: i.e.
the likelihood function of this model assigns a small but equal likelihood to any of the possible behaviors, including the ones facilitated by other models. 
This can be viewed as a model belonging to a random agent. 
We assume that the human, by default,
assigns smaller priors to $\mathcal{M}^0$ than other models. We can now define the following problem:
\begin{definition}
A {\em Generalized Human-Aware Planning Problem} (G-HAP) is a tuple $\Pi_{\mathcal{H}} = \langle \mathcal{M}^R, \mathbb{M}^R_h, P^0_h,$ $P_{\ell}, C_{\mathcal{H}}\rangle$, where $P^0_h$ is the human's initial prior over the models in the hypothesis set $\mathbb{M}^R_h$ and $C_{\mathcal{H}}$ is a generalized cost function that depends on the exact objective of the agent. 
\end{definition}
A solution to G-HAP consists of a behavior that is valid in $\mathcal{M}^R$ and minimizes $C_{\mathcal{H}}$.
In the most general setting, $C_{\mathcal{H}}$ would be a mapping from entire behavior to a cost.
Though internally $C_{\mathcal{H}}$ may be a function that takes into account each of the intermediate steps (not just in $\mathcal{M}^R$ but also the other models in $\mathbb{M}^R_h$). 
While the exact form would depend on the specific agent objectives, in general the cost function may need to consider (1) costs of the action in the sequence in $\mathcal{M}^R$ and their counterparts in each of the models in $\mathbb{M}^R_h$ (2) the state induced by the action in each model (3) possible completions at each intermediate step and their relation to the actual behavior and (4) the beliefs over $\mathbb{M}^R_h$ it may induce. Rather than investigate the space of all possible cost functions, we will ground the discussion by focusing on scenarios and objectives previously studied in the literature.
Throughout the discussion we will use the notations $\tau_{pre}^{i}$ and ${\tau}^{i}_{post}$ for a complete behavior $\tau$ to represent the behavior prefix that would have been observed and the behavior postfix remaining to be executed for a timestep $i$ respectively.
 We will see how this specialization of the framework, naturally gives rise to the specific interpretability measures.
Throughout the discussion we will use the notations $\tau_{pre}^{i}$ and ${\tau}^{i}_{post}$ for a complete behavior $\tau$ to represent the behavior prefix that would have been observed and the behavior postfix remaining to be executed for a timestep $i$ respectively.

% \subsection{Generalized Interpretability}
% \label{revised}
% In the last section we discussed how the observation of robot behavior leads to the human 
% updating both their beliefs about the agent model and possible future actions 
% of the agent. Depending on the objectives the robot may be required to engender specific beliefs on the human's end thereby giving raise to very different behavior on the robot. In this section, we will look at some specific classes of cost functions and how that give rise to existing interpretability measures.

\subsection{Explicability} 
We will start by looking at cases where the agent wants to avoid behaviors that may confuse the observer about the agent model. That is the human should be able to explain the observed behavior with the explicit models they hold. We will refer to such behaviors as {\em explicable behaviors}.
We can capture the generation of such behavior within our framework by using 
a cost function that is proportional to the posterior probability associated with model $\mathcal{M}^0$, i.e.,
\begin{align}
   C_{\mathcal{H}}(\tau) \propto  \sum_{i} \alpha_{i} P(\mathcal{M}^0|\tau_{pre}^{i}) 
\end{align}
Where $\alpha_i \geq 0$ is the weight associated with each timestep $i$. This means the formulation would prefer behavior with high likelihood in the explicit models for timesteps with non-zero weight.
 We will define the explicability score ($\mathcal{E}$) associated with a behavior prefix ($\tau_{pre}$) to be directly proportional to one minus this probability, i.e.,
\begin{align}
\mathcal{E}(\tau_{pre}^{i}) \propto \Sigma_{\mathcal{M} \in \mathbb{M}^R_h\setminus\{\mathcal{M}^0\}} P(\mathcal{M}|\tau_{pre}^{i}) 
\label{exp_gen}
\end{align}
Likelihood functions that assign high probabilities to optimal (or low cost traces), give rise to traces like P1 and P2 in Figure \ref{fig:office-1}, since they correspond to optimal plans in the explicit models considered in the example (i.e. the model for delivering coffee or delivering mail).

\noindent\textbf{Reduction to Previous Explicability Definitions:} 
Previous works generally identify a behavior to be explicable if it meets the human's expectation 
from the agent for the given task \cite{zhang2017plan}. 
In the binary form this is usually taken to mean that a plan is explicable if it is one of the plans that the human expects from the agent \cite{balancing}. 
In the more general continuous form, 
this expectation is taken to be proportional 
to the distance between the observed trace and the
closest expected behavior \cite{kulkarni2019explicable,zhang2017plan}: 
\begin{align}
\tau^{*}_{\mathcal{E}} = \argmin_{\tau} \delta(\tau, \tau^{E}_{\mathcal{M}^R_h})
\end{align}
where $\delta$ is some distance function between two plans and $\tau^{E}_{\mathcal{M}^R_h}$ is the closest expected behavior for the model $\mathcal{M}^R$. While there is no consensus on the distance function or expected behavior, a reasonable possibility for the expected set is the set of optimal 
plans \cite{explain} and the distance can be the cost difference \cite{kulkarni2020designing}.

To see how our framework subsumes earlier works, lets start by plugging in the two assumptions made by the original works, namely (1) the human only has one explicit model about the agent (i.e. $\mathbb{M}^R_h = \{\mathcal{M}^R_h, \mathcal{M}^0\}$) and (2) the explicability is measured over the entire plan (i.e. $\alpha_i=0$ for all $i$ other than the last step). Thus the cost function is dependent only on the explicability  of the entire behavior
\begin{align}
\mathcal{E}(\tau) \propto  P(\mathcal{M}^R_h|\tau)
\propto  P(\tau|\mathcal{M}^R_h) * P(\mathcal{M}^R_h)
\end{align}
Since the observed prefix is the entire plan, we can directly use the likelihood function
\begin{align}
\mathcal{E} (\tau) \propto  P_{\ell}(\mathcal{M}^R_h, \tau) * P(\mathcal{M}^R_h)
\end{align}
Let us consider two plausible likelihood models. 
First, for a normative model where the agent is expected to be optimal,
$P_{\ell}(\mathcal{M}^R_h, \tau_{pre})$ 
assigns high but equal probability to all the optimal plans and $0$ probabilities for the others.
%is either $\frac{1}{m}$ ($m$ being the number of optimal plans) leading to high explicability or $0$ for not explicable. 
This is the original binary explicability formulation used by \cite{balancing,chakraborti2019explicability}. 

Another possible likelihood function is a noisy rational model  \cite{fisac2020generating} given by:
\begin{align}
P_{\ell}(\mathcal{M}^R_h, \tau) \propto e^{ - \beta \times C(\tau)} \propto e^{ \beta \times C(\tau^{*}) - C(\tau)}
\label{equ:exp}
\end{align}
%\note{put $\beta$ in}
where $\tau^{*}$ is an optimal behavior in $\mathcal{M}^R_h$, $C(\tau) \geq C(\tau^{*}) \geq 0$ for $\mathcal{M}^R_h$.
This maps the formulation to the distance based definition
as in \cite{kulkarni2020designing} where a cost-based distance is defined. 
We can also recover the earlier normative model by setting $\beta \rightarrow \infty $ and model $\mathcal{M}^0$ by setting $\beta=0$.

Going back to the original motivation of explicability, it was meant to capture the user's understanding of the agent behavior generation process (which includes both its perceived model and computational component). 
Earlier formulations rely on using the space of expected plans 
as a proxy of this process. 
This is further supported by the fact that the works that have looked at updating the
human's perceived explicability value of a plan do so by providing information 
about the model and not by directly modifying the human's understanding of the 
expected set of plans \cite{explain}. Thus our formulation of explicability directly in terms of the human's beliefs about the agent's model connects to the original motivation of explicability definitions.\looseness=-1

\noindent\textbf{Novel Properties of Generalized Explicability:}
An interesting side-effect of a probability-based explicability formulation 
is that, 
the probability of behavior and hence the explicability score can now be affected by the presence or absence of other plans. For example, consider two scenarios, 
one where $\mathbb{M}^R_h$ contains $\mathcal{M}_1$ and $\mathcal{M}^0$ and 
another where it contains $\mathcal{M}_2$ and $\mathcal{M}^0$. 
Now consider a behavior trace $\tau$ such that it is equidistant 
from an optimal plan in both models $\mathcal{M}_1$ and $\mathcal{M}_2$. 
Even though they are at the same distance, the trace may be more explicable in the first scenario than in the second, if the second scenario allows for more traces that are closer. Assuming the probability of choosing optimal plans isn't reduced, introducing new plans into the sample space better than the current trace would cause more probability to be assigned to those and thus less to the trace in question. 
We argue that this makes intuitive sense for explicability since the user should be more surprised in the second scenario as the agent would have ignored many more behaviors that the observer would have considered desirable.

\noindent \textbf{Property 1} \  \textit{Explicability of a trace is dependent 
not only on the distance from the expected plans but also on the presence or 
absence of plans close to the expected plans.}

Here the plan remains explicable whether or not the observation leads to all the probability being assigned to a single model versus being distributed across multiple models.
This means that the formulation doesn't require the human to have a single explanation for the behavior, rather it allows their belief to be distributed across multiple hypotheses.
While the exact values would depend on the likelihood function, in the office robot scenario our formulation would assign high explicability scores (need not be the same) to both $P1$ and $P2$. 
For $P1$, the probability mass would be distributed across the two possible hypotheses corresponding to the two goals, while for $P2$ the probability mass is centered around the model corresponding to the goal to fetch coffee.  

\noindent\textbf{Property 2} \textit{Explicability is agnostic to whether it is supported by multiple models or by a single one.}

Further, the explicability of a trace is now controlled by the priors on the models. E.g., a trace that is only possible in a model with low prior will not have high explicability score even if it is highly likely in that model.

\subsection{Legibility}
The next class of behavior is the one where the agent is trying to choose behavior that increases the agent's belief about some component (captured by the parameter $\theta$) of the agent model. Such behavior could be especially important when the achievement of the agent's objective is tied directly to that model component. That is the agent can only achieve the desired outcome if its model possesses that parameter value. An obvious example would be establishing if the end-goal itself is what the human desires, but this could also be in relation to other model parameters. Thus inducing high confidence in relation to such model parameters in the human's mind could be tied intimately with engendering trust in the human that the agent can achieve the desired objectives.
\begin{align}
   C_{\mathcal{H}}(\tau) \propto \sum_{i} \alpha_{i} *( 1 - P(\theta = \theta(\mathcal{M}^R)|\tau_{pre}^{i}))
\end{align}
Keeping with the existing literature, we will refer to such behaviors as {\em Legible Behavior}, with the actual legibility score of a behavior prefix being proportional to the probability of the parameter being the true value
\begin{align}
\mathcal{L}^{\theta}(\tau_{pre}) \propto P(\theta = \theta(\mathcal{M}^R)|\tau_{pre}) \\\propto \Sigma_{\mathcal{M} \in \mathbb{M}^R_h\setminus\{\mathcal{M}^0\} ~\textrm{Where}~ \theta(\mathcal{M}^R)=\theta(\mathcal{M})} P(\mathcal{M}|\tau_{pre})  
\label{equ:leg}
\end{align}
We skip $\mathcal{M}^0$ since it doesn't correspond to an explicit model in the human's mind. In the context of Figure \ref{fig:office-1}, a plan prefix with high legibility score for the goal of deliver coffee would be P2 as compared to the other options illustrated. Since P1, allows for an optimal completion for both objective and P3's completions in both models are equally bad. As we will see, while original formulations might assign P4 as a more legible option given the fact that it would assign zero probability to delivering mail, our formulation allows for the possibility that P4 may lead to more probability getting assigned to $\mathcal{M}^0$.

\noindent\textbf{Reduction to Previous Legibility Definitions:} 
Legibility was originally formalized 
\cite{dragan2013legibility}
as the ability of a behavior to reveal its underlying objective. 
This involves a human who is considering a set of possible goals ($\mathbb{G}$) of the agent and is trying to identify the real goal 
by observing its behavior. 
Legibility is, thus, the 
maximization of the probability of the real goal
through behavior:
\begin{align}
\hat{\tau}^{*}_{\mathcal{L}} = \argmax_{\tau_{pre}} P(G^R| \tau_{pre})
\end{align}
where $G^R$ is the agent's true goal.
While originally introduced in the context of motion planning, 
this was later adapted to task planning by
\cite{macnally2018action}, and 
generalized to implicit communication of beliefs 
when the human has partial observability by \cite{kulkarni2019unified} 
as well as to implicit communication of any model parameter by \cite{SZ:MZaamas20}.

To keep the discussion in line with previous works, we will focus our attention on communicating end-goals (over arbitrary parameters). 
Some central assumptions made by earlier works is that the model only differs in terms of the end goal and the actual model is part of the set ($\mathcal{M}^R \in \mathbb{M}^R_h $). Also, the agent is expected to communicate its information as early as possible, so earlier $\alpha_i$ terms are given higher weights than the latter ones. They also assume that at no point would the human consider goals outside the explicit ones she had in mind. That is the possibility that she may be wrong about the original model and that the agent may be possibly trying to achieve something she didn't consider before would never cross her mind. In our framework, this would correspond to assigning a zero prior to $\mathcal{M}^0$. Thus the legibility score here would be
\begin{align}
\mathcal{L}^{\theta}(\tau_{pre}) \propto P(M^R|\tau_{pre})  
\end{align}
A zero prior on $\mathcal{M}^0$ means the agent can create extremely circuitous routes as legible behavior provided the behavior is more likely in the agent model than others. This means that regardless of how suboptimal the plan is in the agent model (or ones with the parameter value), given its even lower probability in other models (or for other parameter values) the agent model will get assigned higher posterior probability and thus higher legibility score.
For example in Figure \ref{fig:office-1}, the restricted formulation would select the prefix $P4$ highlighted in red
in order to reveal the goal of delivering coffee, eventhough that corresponds to an extremely
sub-optimal plan given the set of possible plans.

\noindent\textbf{Novel Properties of Generalized Legibility:}
A core assumption relaxed by the general formulation is that we now allow for the possibility that the human could be surprised by unexpected behavior and they may form new hypotheses about the agent. If you assume a non-zero prior for $\mathcal{M}^0$, then in cases where the agent presents an extremely suboptimal behavior they have a new hypothesis they can consider. That is they can now shift some of their belief to the fact that they may have been originally wrong  about the agent model. Going back to the case of route $P_4$ in Figure \ref{fig:office-1}, given how far it is from the optimal any completion of that prefix would have extremely low likelihood in the model for delivering coffee as opposed to $\mathcal{M}^0$ where that path is as likely as any other. This means our formulation now assigns more weight to $\mathcal{M}^0$ and thus capturing the fact that when presented with highly unlikely behavior, the observer may question their beliefs about the agent. Which brings us to the property

\noindent\textbf{Property 3} \textit{Inexplicable plans are also illegible.}

We believe allowing for such uncertainty is essential to capture more realistic human-robot interaction as it is rare for people to have absolute certainty about the agent models (and even discard the possibility that something might have just gone wrong with the agent). Also if we wish to move to a more longitudinal setting, indicating that the human no longer believes in one of the possible hypotheses in the set may not be enough, but we may need to explicitly try to identify what the newly formed hypothesis might be.

\subsection{Predictability}
The final case is one where the agent is interested in communicating to the human the future behavior it will be selecting. In this case, the agent would be required to choose behavior prefixes that allow the human to correctly guess the rest of the behavior the agent will follow with high confidence. This may be useful in cases where the agent may be sharing a workspace with the observer and may want to allow the observer to take into account future agent actions when coming up with their plans.
\begin{align}
   C_{\mathcal{H}} (\tau) \propto \sum_{i} \alpha_{i}*(1 - P({\tau}^{i}_{post}|\hat{\tau}_{pre}^{i}))
\end{align}
This gives us {\em predictable behavior}. Further, $P({\tau}^{i}_{post}|\hat{\tau}_{pre}^{i})$ denotes the predictability score for the prefix $\tau_{pre}^{i}$ (with respect to the completion ${\tau}^{i}_{post}$) 
\begin{align}
\begin{split}
\mathcal{P}^{{\tau}^{i}_{post}} (\tau_{pre}^{i}) \propto  P({\tau}^{i}_{post}|\tau_{pre}^{i})\\
\propto  \sum_{\mathcal{M} \in \mathbb{M}^R_h} P({\tau}^{i}_{post}|\tau_{pre}^{i}, \mathcal{M}) \times P(\mathcal{M})    
\end{split}
\end{align}
From Figure \ref{fig:office-1}, a plan prefix with high predictability would be $P3$, since both of the goals have high prior the only possible completion corresponds to the same path (assuming no separate action for picking up of items). Thus that completion has high probability and hence high predictability.\looseness=-1

\noindent\textbf{Reduction to Previous Predictability Definitions:} We need to incorporate two main assumptions into the framework to reduce it to existing definitions of predictability: (1) the human observer only has a single explicit model about the agent and this is equal to the actual agent model $\mathbb{M}^R_h = \{\mathcal{M}^R, \mathcal{M}^0\}$ and (2) the user will not form new hypothesis about the agent regardless of how unexpected the behavior is (i.e. the $\mathcal{M}^0$ prior is zero). Thus we get:
\begin{align}
    \mathcal{P}^{\tau} (\tau_{pre}) \propto  P(\tau'=\tau |\tau_{pre}, \mathcal{M}^R)
\end{align}
This directly maps to the predictability measure as defined in earlier works \cite{fisac2020generating}.
Previous works have also looked at the possibility of generating $k$ step predictable plans, i.e., plans that try to guarantee predictability only after $k$ steps. This allows for the system to 
choose unlikely prefixes for cases where the agent is only required to achieve required levels of predictability after the first $k$ steps. We can capture such optimization preferences by setting $\alpha_i$ for all but $i=k$ to zero. Going back to the example, prefix $P3$ optimizes for predictability for $k=5$.  

\noindent\textbf{Generalized Predictability:}
Our generalization introduces two new aspects to the predictability formulation. The fact that the human now considers potential models and we also introduce the new hypothesis $\mathcal{M}^0$. However, the formulation marginalizes out the model and thus effectively for a given prefix the human observer has to consider all the possible completions of the prefix in each of the individual models. Thus even if the trace is perfectly predictable in an individual model, the fact that the human has uncertainty over the models means the prefix may not be predictable. On the other hand, the fact that $\mathcal{M}^0$ assigns equal probability to all the possible completions would mean that the introduction of this new hypothesis would have less of an influence on the resulting predictability score.

\subsection{Deception and Interpretability}
\label{adversarial}

The interpretability measures being discussed involve leveraging reasoning processes at the human's end to allow them to reach specific conclusions. 
At least for legibility and predictability, the behavior is said to exhibit a 
particular interpretability property only when the conclusion lines up with 
the ground truth at the agent's end. But as far as the human is concerned,
they would not be able to distinguish between cases where the behavior is 
driving them to true conclusions or not. 
This means that the mechanisms used for interpretability could be easily leveraged to perform behaviors that may be adversarial \cite{chakraborti2019explicability}.
Two common classes of such behaviors are deception and obfuscation. 
Deceptive behavior corresponds to behavior meant to convince the user of incorrect information about the agent model or its
future plans \cite{masters2017deceptive}:
% while the latter is meant to confuse the user \note{TC: cite}. 
% This might be equally applicable to both model information and future plans.
\begin{align}
\mathcal{D}^{\mathbb{M}^R_h}(\tau_{pre}) \propto  -1 * P(\mathcal{M}^R|\tau_{pre})
\end{align}

% This formulation can also be adapted to parameters and plans.

Adversarial behaviors meant to confuse the user
are either inexplicable plans that increase the posterior
on $\mathcal{M}^0$ or, plans that 
actively obfuscate \cite{keren2016privacy,kulkarni2019unified}:
\begin{align}
\mathcal{O}^{\mathbb{M}^R_h}(\tau_{pre}) \propto H(\mathbb{M}^R_h|\tau_{pre})
\end{align}

This is proportional to the conditional entropy of the model distribution given the observed behavior.

% This formulation can also be adapted to parameters and plans.

With explicability, the question of deceptive behavior becomes 
interesting, since explicable plan generation is relevant when the actual agent model may not be part of the human's expected set of models (else the agent could just follow its optimal behavior). By choosing to generate plans that align with a non-true model, explicability can be seen as deceptive behavior as it is reinforcing incorrect notions about the agent's model. Such plans would have a high deceptive score per the 
formulation above (since $P(\mathcal{M}^R|\tau) = 0$). 
One can argue that explicable behaviors are white lies in such scenarios as the goal here is just to ease the interaction and the behavior is not driven by any malicious intent. One could even further restrict the 
explicability formulation to a version that only lies by omission
by restricting the agent to just behavior optimal in the original agent 
model. The agent chooses from this set the one that best aligns with the human's expectation. It is a lie by omission in the sense that while the agent has not explicitly been deceptive, by choosing behavior that aligns with human's expectations, it is maintaining the human's incorrect beliefs.

\vspace{-10pt}
\section{Conclusion}
\label{conclusion}

% Most works on interpretable behavior generation have focused on 
% studying clean mathematical models that reflect our intuitions
% about desirable behavioral properties. However, by focusing 
% on individual behavioral properties, we overlook interactions that are important for 
% successful deployment of such methods in real-world applications. 
% In this work, we introduced a framework that is able to synthesize and present possible generalizations for existing works within the context of interpretable behavior generation. 
% We will now discuss a few more implications of 
% the proposed formulation and directions for future work.
Works on interpretable behavior generation have generally focused on studying and defining individual human-aware measures under limited settings. We hope that by placing and studying human-aware planning in a more general context we are able to not only see connections between these works that were previously ignored but also help formalize new colloboration strategies. Below we have provided a brief discussion of some additional implications of the framework.\\
%\textbf{Legibility, Explicability, and Expectation of the Agent Modeling the Human:}
\textbf{Legibility, Explicability}
These notions are related to the human's desire to recognize the model \cite{model-rec}). Our formulation shows that outside
limited cases, legibility, and explicability are closely connected. 
Earlier works have been separating these measures by assuming away either legibility, like in existing explicability works with the human's hypothesis consisting of a single model \cite{zhang2017plan,kulkarni2019explicable}, 
or by assuming away explicability by assigning zero prior on $\mathcal{M}^0$ for legibility \cite{dragan2013legibility,dragan2013generating,macnally2018action,kulkarni2019unified,SZ:MZaamas20}. 
Interestingly, in cases where the human 
is aware that the agent is trying to be legible or more generally they know the agent is trying to model the observer, the human may be more open to suboptimal behavior from the agent as they might attribute it to trying to communicate.
However, this does not eliminate $\mathcal{M}^0$
but instead introduces a new level of nesting for reasoning. This comes with all the known complexities and pitfalls of reasoning with nested beliefs \cite{fagin2003reasoning}. Though studying a limited amount of additional nesting could be important especially in cases where the agent plans to leverage communication. Since communication strategies make the most sense when the human is expecting the agent to model them.\looseness=-1\\
% \textbf{Planning:} 
% The next logical step for this work
% would be to be able to generate plans with the metrics
% in the unified framework. A good starting point may be to 
% compile the problem to a classical planning problem,
% as done in \cite{exact} for explicable plans. Though here we will need to replace the specialized reasoning they perform for explicability with the more general mental model reasoning performed in our framework.\\
\textbf{Longitudinal Interactions:}
Our formulation currently looks at interpretability metrics for 
one-off interactions only. 
In cases where a human interacts with the agent for a long period, 
we can expect the user to start with a uniform distribution over
models and a low probability for $\mathcal{M}^0$. 
In order to take a more long-term view of the human's interaction 
with the same agent (say, over a time horizon), 
legibility and predictability measures can be handled by directly 
carrying over the posterior from each interaction to the next one. 
However, for explicability more care needs to be taken. 
For example, \cite{kulkarni2020designing} hypothesize 
a possible discounting of inexplicable behavior. 
They argue that after the first inexplicability, a human would
be less surprised when similar inexplicable behavior was again
presented to her. 
Part of this discounting can be explained by the human forming new hypothesis that explain the unexpected behavior and using that to analyze future agent behavior. So as mentioned earlier, going to a longitudinal setting may require introducing new mechanisms to identify such newly formed hypothesis.

\section*{Acknowledgements}
We would like to thank Dr. Tathagata Chakraborti for his contributions to an earlier version of this paper and for extensive discussions on the topic. This research is supported in part by ONR grants N00014-16-1-2892, N00014-18-1- 2442, N00014-18-1-2840, N00014-9-1-2119, AFOSR grant FA9550-18-1-0067, DARPA SAIL-ON grant W911NF-19- 2-0006, NASA grant NNX17AD06G, and a JP Morgan AI Faculty Research grant.
\bibliographystyle{named}
\bibliography{aaai_subm}

\end{document}